\documentclass[letterpaper]{article} 
\usepackage{aaai24}  
\usepackage{times}  
\usepackage{helvet}  
\usepackage{courier}  
\usepackage[hyphens]{url}  
\usepackage{graphicx} 
\usepackage{natbib}  
\usepackage{caption} 
\usepackage{algorithm}
\usepackage{algorithmic}

\usepackage{booktabs}
\usepackage{multirow}
\usepackage{amssymb}
\usepackage{amsmath}

\usepackage{newfloat}
\usepackage{listings}
\DeclareCaptionStyle{ruled}{labelfont=normalfont,labelsep=colon,strut=off} 
\floatstyle{ruled}
\newfloat{listing}{tb}{lst}{}
\floatname{listing}{Listing}
\title{ Generative Model-based Feature Knowledge Distillation for Action Recognition}
\author{
    Guiqin Wang\textsuperscript{\rm 1,\rm 3}
\quad Peng Zhao\textsuperscript{\rm 1,\rm 3}\thanks{Corresponding author: Peng Zhao (p.zhao@xjtu.edu.cn)}
\quad Yanjiang Shi\textsuperscript{\rm 1,\rm 3}
\quad Cong Zhao\textsuperscript{\rm 2, \rm 3}
\quad Shusen Yang\textsuperscript{\rm 2, \rm 3}
}
\affiliations{
    \textsuperscript{\rm 1} School of Computer Science and Technology, Xi'an Jiaotong University\\
\textsuperscript{\rm 2} School of Mathematics and Statistics, Xi'an Jiaotong University\\
\textsuperscript{\rm 3} National Engineering Laboratory for Big Data Analytics, Xi'an Jiaotong University

}

\usepackage{bibentry}

\begin{document}

\maketitle

\begin{abstract}
Knowledge distillation (KD), a technique widely employed in computer vision, has emerged as a de facto standard for improving the performance of small neural networks. However, prevailing KD-based approaches in video tasks primarily focus on designing loss functions and fusing cross-modal information. This overlooks the spatial-temporal feature semantics, resulting in limited advancements in model compression. Addressing this gap, our paper introduces an innovative knowledge distillation framework, with the generative model for training a lightweight student model. In particular, the framework is organized into two steps: the initial phase is Feature Representation, wherein a generative model-based attention module is trained to represent feature semantics; Subsequently, the Generative-based Feature Distillation phase encompasses both Generative Distillation and Attention Distillation, with the objective of transferring attention-based feature semantics with the generative model. The efficacy of our approach is demonstrated through comprehensive experiments on diverse popular datasets, proving considerable enhancements in video action recognition task. Moreover, the effectiveness of our proposed framework is validated in the context of more intricate video action detection task. Our code is available at \url{https://github.com/aaai-24/Generative-based-KD}.


\end{abstract}
\section{Introduction}
In recent years, various deep learning technologies have achieved significant success in the domain of intelligent video analysis~\cite{foo2023system,yang2022massive}. Specifically, action recognition stands as a pivotal task within intelligent video analysis, entailing the categorization of action instances into corresponding labels. Recently, substantial enhancements have been achieved in the performance of action recognition~\cite{sun2022human}. Intuitively, a larger backbone model often corresponds to improved performance. This perspective has prompted numerous researchers to devise intricate backbone models for capturing video-centric feature information(\textit{e.g.}, C3D~\cite{xu2017r}, I3D~\cite{carreira2017quo}, S3D~\cite{xie2018rethinking}). These efforts have yielded commendable results in the field of video action recognition. However, the deployment of a larger backbone introduces extremely high resource and memory constraints, making it impractical for edge devices with limited resources. To address this challenge, Hinton et al. proposed Knowledge Distillation (KD) as a solution, which facilitates the transfer of learned knowledge from a heavyweight model (teacher model) to a lightweight model (student model).


In accordance with the fundamental objective of model compression, KD methodologies primarily encompass two investigational paradigms: logits-based KD and feature-based KD. Logits-based KD~\cite{hinton2015distilling,li2023curriculum} entails the condensation of a large teacher model into a more compact student counterpart, achieved through the conveyance of "dark knowledge" via soft labels generated from the teacher's output. Following the introduction of the approach by~\citeauthor{romero2014fitnets} in 2015, comparative assessments have consistently validated the superiority of feature-based KD methodologies~\cite{xu2020feature,zhang2020improve,yang2022masked} across a diverse array of tasks~\cite{zhao2022decoupled}. Consequently, scholarly attention has prominently shifted towards the extraction of knowledge from intricate features residing within intermediary layers of model.


\begin{figure}[t]
  \centering
  \includegraphics[width=0.48\textwidth]{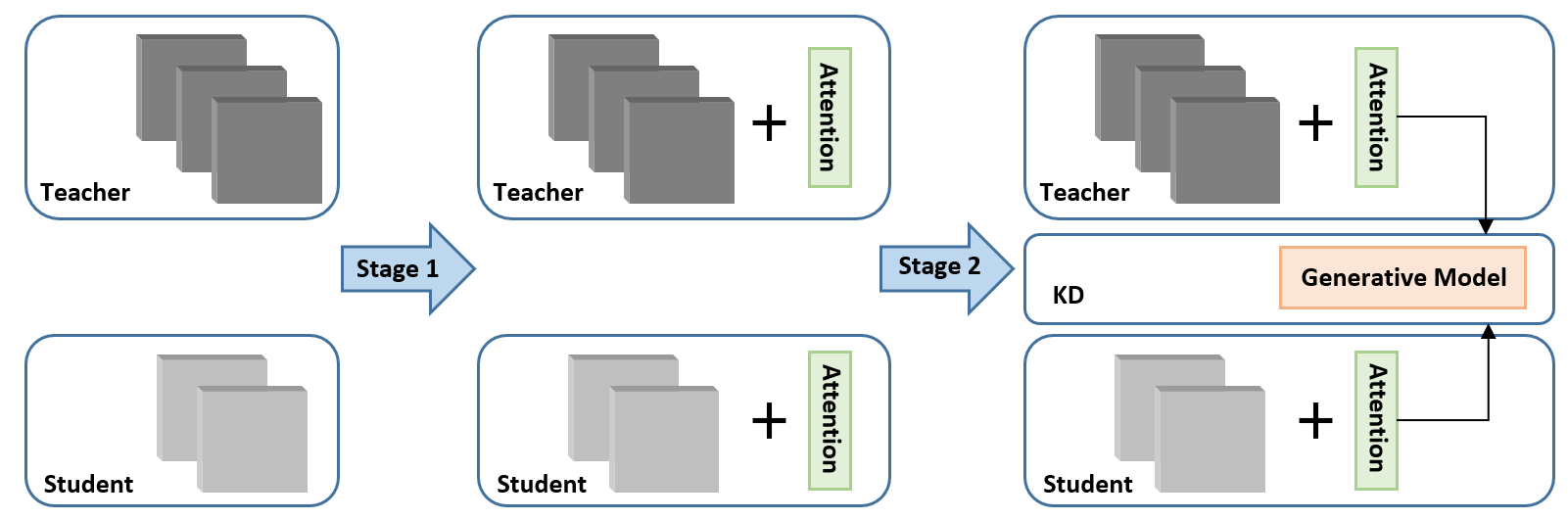}
  \caption{Our proposed framework includes two stages: Stage $1$ adds an attention module to represent feature semantics; Stage $2$ builds a generative-based KD module to distill feature knowledge from teacher model.}
  \label{fig:intro 1}
\end{figure}

However, the majority of feature-based KD methods~\cite{zhao2022decoupled,zhang2020improve,xu2020feature} are centered on the design of diverse loss functions to match feature maps, disregarding the inherent significance of variational feature semantics. In fact, particularly in the context of videos, features inherently contain intense semantic information~\cite{quader2020weight}, primarily stemming from both temporal and spatial variations. These feature semantics play a crucial role in enhancing the precision of the student model by transferring the variations of features from the teacher model. Regrettably, prevailing solutions often overlook such semantics, resulting in trivial improvements in the context of video tasks. Consequently, current KD methodologies predominantly cater to image-related tasks~\cite{zhao2022decoupled,yang2022focal,Lin_2022_CVPR}. Within the domain of video action recognition, the predominant focus of KD methodologies centers on cross-modal distillation, aimed at elevating the accuracy of models~\cite{liu2021semantics,thoker2019cross,dai2021learning}. Therefore, a noticeable research gap emerges in the realm of KD-based methods conducive for compressing 3D-CNN models, particularly within the context of video action analysis.

To address these issues, we proposed a novel generative model-based feature knowledge distillation framework for video action recognition. This framework enables the efficient transfer of feature semantics from heavyweight models (teacher model) to lightweight models (student model) between intermediate layers. Specifically, our proposed approach consists of two components. Firstly, we designed a Feature Representation module that leverages a generative-based attention model to acquire feature semantics within the 3D-CNN architecture. Secondly, we built a generative-based KD module, encompassing both Generative Distillation and Attention Distillation processes. This module is designed to distill attention-based feature information from the teacher model, as illustrated in Fig.~\ref{fig:intro 1}. The Generative Distillation component is tailored to optimize the Feature Representation module of the student model by matching reconstructed features, thereby facilitating the learning of the generative-based attention model. Subsequently, the Attention Distillation process is orchestrated to transfer attention-based feature semantics by matching attention maps, conditioned on unchanged attention-based feature distributions.

To our best knowledge, we are the first to consider the temporal variation of feature semantics and study the generative model mechanism in KD. We design a generative model-based KD framework that effectively enhances the compression performance of 3D-CNN models. Our main contributions are summarized as follows:

\begin{itemize}

    \item To distill temporal-spatial feature information, we design a novel attention module that leverages the generative model to represent feature semantics within the 3D-CNN architecture.
    
    \item We build a new framework that firstly introduces the novel concept of utilizing a generative model for distilling attention-based feature knowledge. Particularly, our KD framework is the first to compress 3D-CNNs on video, with a generative model to transfer the temporal-spatial information.
       
    \item Based on extensive experiments, our approach demonstrates remarkable performance improvements across various network architectures on two prominent action recognition datasets. Additionally, we extend our framework to more complex task, action detection, which also provides considerable performance enhancements.
    
\end{itemize}

\section{Related Work}\label{sec:Related work}

\subsection{Action Recognition}
Most previous work of CNN architectures on action recognition can be categorized into two groups: 3D CNNs~\cite{carreira2017quo,xu2017r,ji20123d,hara2017learning} and partial 3D CNNs~\cite{tran2018closer,qiu2017learning,xie2018rethinking}. 3D CNNs were first proposed in~\cite{carreira2017quo,tran2015learning}, which consider a video as a stack of frames to learn spatiotemporal features of action by 3D convolution kernels. Furthermore, \citeauthor{carreira2017quo} proposed I3D~\cite{carreira2017quo} to capture spatial and temporal information by fusing RGB and optical flow based on the 3D convolution. Partial 3D CNNs had been proposed in~\cite{tran2018closer, qiu2017learning}, which replaced 3D convolutions with depth-wise separable convolutions to reduce resource cost. Meanwhile, \citeauthor{xie2018rethinking} replaced 3D convolutions with 2D convolutions to reduce computational complexity. 
However, 3D CNNs consume a substantial amount of resources to achieve high accuracy, which makes them unsuitable for deployment on resource-constrained devices. While the utilization of Partial 3D CNNs mitigates the resource demands, they still exhibit an accuracy gap when compared to full-fledged 3D CNNs.

\subsection{Generative Model}
Generative model~\cite{goodfellow2014generative,kingma2013auto} develops rapidly in recent years by combining with deep learning. GAN~\cite{goodfellow2014generative} maximises the approximate real data distribution information between a subset of the generating variables and the output of a recognition network. However, GAN learns distribution implicitly and lacks of sample diversity. VAE~\cite{kingma2013auto} approximates the real distribution by optimizing the variational lower bound on the marginal likelihood of data. 
However, VAE is not suitable to model the distribution of multi-modal output~\cite{sohn2015learning}. In order to transfer attention to guide knowledge distillation, we use conditional VAE~\cite{sohn2015learning} to model the feature semantic distribution conditioned on attention value.

\subsection{Knowledge Distillation}
Knowledge distillation was first proposed in~\cite{hinton2015distilling}, which transfers the output probability distributions via soft labels produced by teacher. Furthermore, \citeauthor{romero2014fitnets} proposed to distill the feature representation from penultimate layer, named feature knowledge distillation. \citeauthor{xu2020feature} proposed feature normalized knowledge distillation to reduce the impact of label noise. Additionally, \citeauthor{zagoruyko2016paying} proposed attention knowledge distillation, which tries to match the attention map to transfer feature knowledge. Concurrently, there are efforts to integrate logits-based methodologies with feature-based approaches~\cite{zhao2020highlight,shen2019customizing}, all with the goal of enhancing overall model performance. However, these approaches primarily concentrate on aligning feature and attention maps, thereby potentially neglecting the underlying feature semantics.

In the context of knowledge distillation applied to action recognition, the majority of endeavors~\cite{crasto2019mars, stroud2020d3d, thoker2019cross} predominantly focus on cross-modal distillation strategies. These strategies aim to enhance model performance by effectuating the transfer of optical flow (teacher model) insights to RGB flow (student model). However, the existing landscape of knowledge distillation-driven frameworks for action recognition is deficient in an established approach to model compression and the reduction of computational overhead.

\section{Method}\label{sec:Method}

In this section, we first introduce the principle of feature KD in action recognition task. Then we present the framework of our KD framework and introduce the two sub-modules.

\subsection{Definition}
\subsubsection{Feature Distillation} The philosophy of knowledge distillation is to train the compact student model to approximate the capability of the cumbersome teacher model. In specific, suppose we have a pre-trained teacher model $T$ and an untrained student model $S$, which are parameterized as neural networks in this work. Let's denote the output feature map $F \in R^{T \times C \times HW}$ of 3D-CNN, where $T$, $C$ and $HW$ represent time, channels, and spatial dimensions, severally. 
For better illustration, we denote $F_T$ and $F_S$ as the feature maps from the layer of the teacher and student model. For feature KD, the distillation distance between student model and teacher model is calculated by the two feature maps:
\begin{equation}
    \mathcal{L}_{KD} = \frac{1}{n} \Vert f(F_T) - f(F_S) \Vert_2^2,
\end{equation}
where $f(\cdot)$ is an explicit mapping function, $n$ is the temporal dimension. The student model is encouraged to minimize the objective function $\mathcal{L}_{KD}$ to mimic the teacher model. Nonetheless, since the feature semantics is neglected in this fashion, the student is not capable of learning the temporal dependence from the teacher in action recognition.

\subsubsection{Feature Representation} For action recognition task, we first define an attention module to represent feature semantics, which also effectively improves recognition performance. We learn a generative model to optimize attention $\lambda$ by leveraging feature $F$ and action classes $C$. For simplifying the problem, we transform the optimization target using Bayes’ theorem:
\begin{equation}\label{equ:att}
\begin{aligned}
   \mathop{\max}_{\lambda \in \left[0,1\right]}{\log p(\lambda \vert F,C)}  & = \max \limits_{\lambda \in \left[0,1\right]} \log p \left(C\rvert F, \lambda \right) + \log p \left(F\rvert \lambda \right) \\
   & + \log p \left(\lambda \right) - \log p \left(F, C, \lambda \right) \\ 
   & \simeq \mathop{\max}_{\lambda \in \left[0,1\right]}{\log p(C \vert F,\lambda)+\log p(\lambda|F)},
\end{aligned}
\end{equation}
in the last step, we discard the constant term $ \log p \left( F, C, \lambda \right) $ and set $ \lambda $ as a uniform distribution. 

As Equ.~\ref{equ:att} means, when we optimize the attention value, we not only use feature semantics and attention to improve the categories performance (the first term), but also ensure the attention-based feature distribution consistent with the original feature distribution (the second term).
\begin{figure*}[t]
  \centering
  \includegraphics[width=0.88\textwidth]{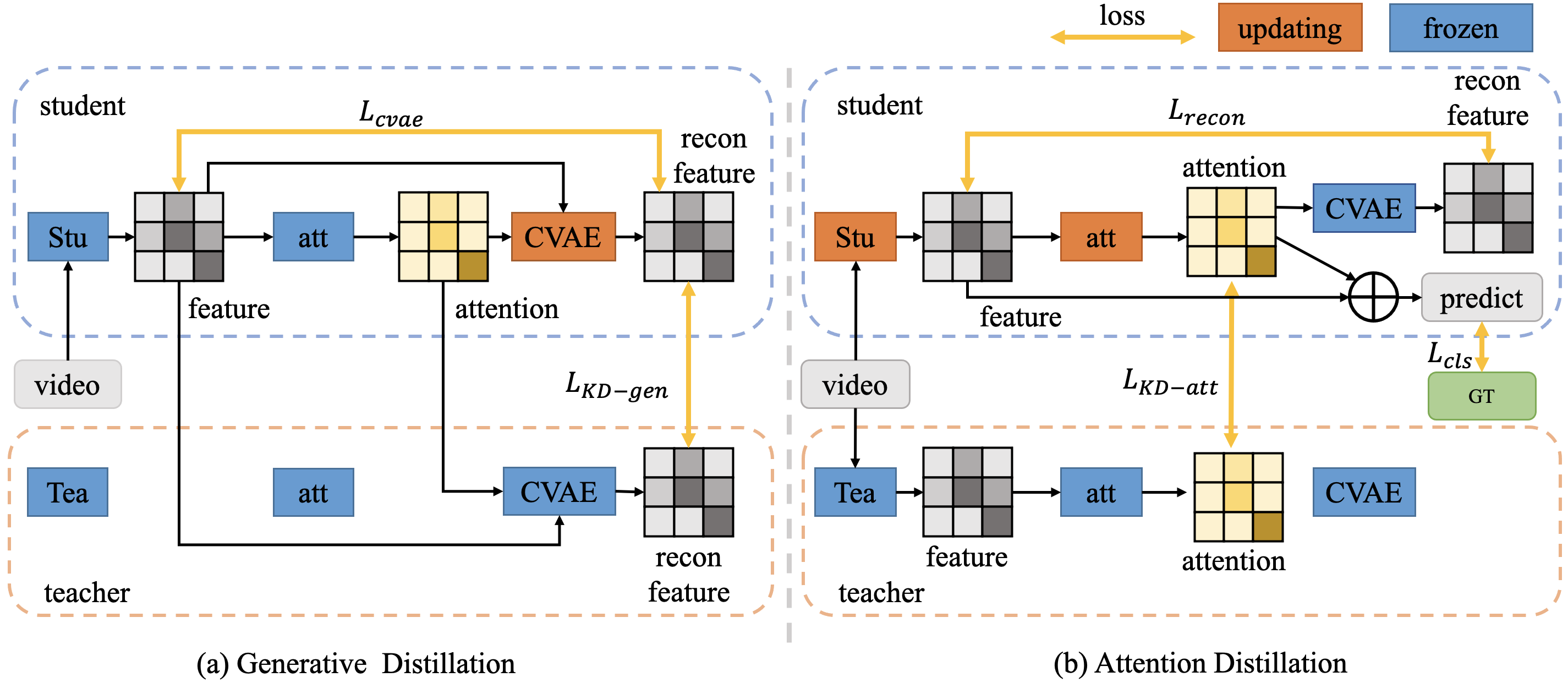}
  \caption{The pipeline of our KD framework. Our KD model is trained alternatively in two stages. In stage $1$, the Generative Distillation is trained with self-reconstruction loss $\mathcal{L}_{CVAE}$ and  generation distillation loss $\mathcal{L}_{KD-gen}$, which is to match reconstructed feature. In stage $2$, the Attention Distillation is updated with representation loss $\mathcal{L}_{recon}$, attention distillation $\mathcal{L}_{KD-att}$ and classification loss $\mathcal{L}_{clf}$, which is to distill the attention-based feature semantics.}
  \label{fig:method 1}
\end{figure*}
\begin{figure}[t]
  \centering
  \includegraphics[width=0.46\textwidth]{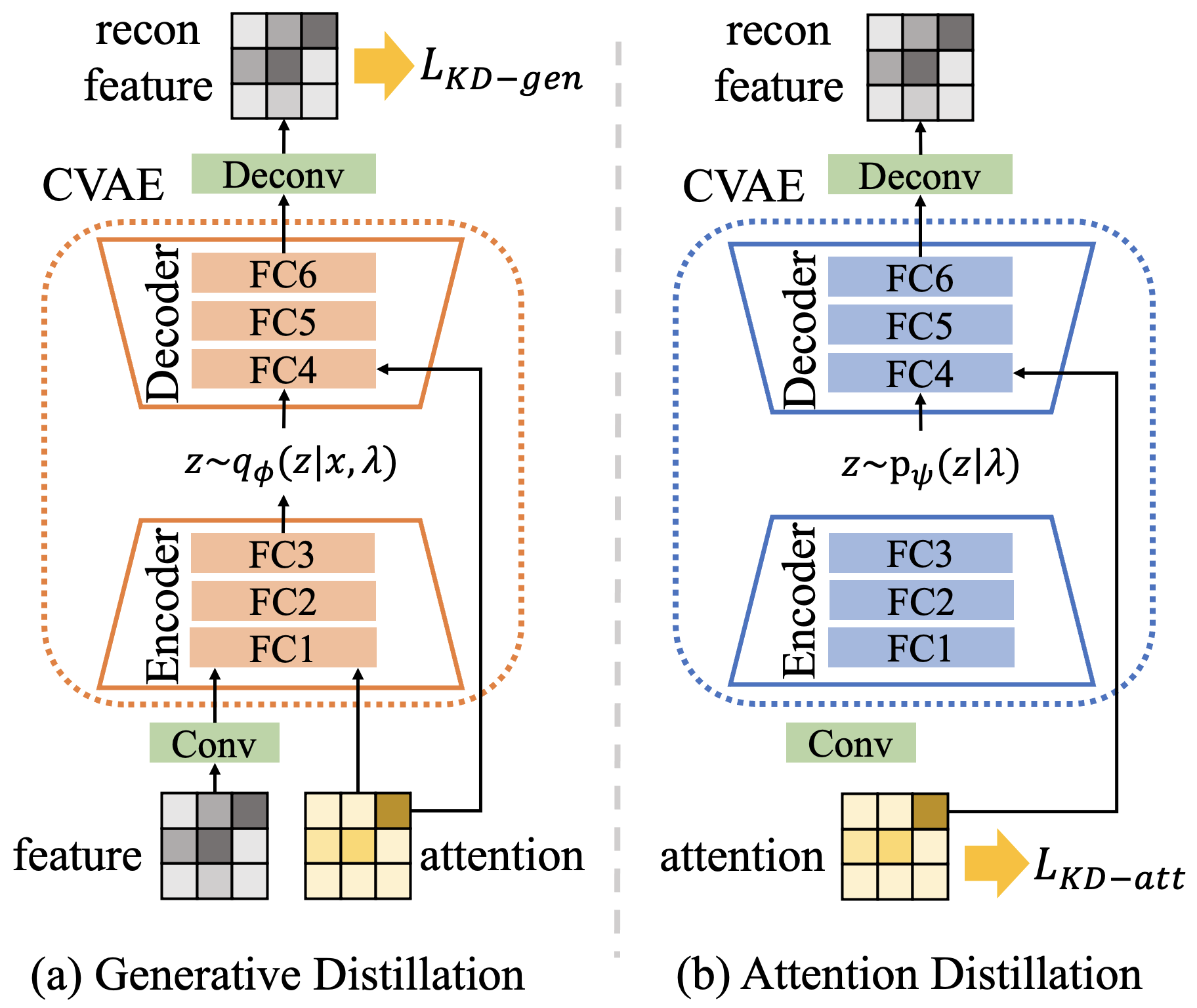}
  \caption{The architecture of our proposed KD framework. It includes the Generative Distillation module and the Attention Distillation module.}
  \label{fig:meth 2}
\end{figure}
\subsection{Feature Knowledge Distillation}

This work employs the feature knowledge distillation to help the student model learn the feature representation. We extract the semantics, expressed as attention map, from a trained teacher model and ask the student model to mimic it.

The pipeline of our KD framework is illustrated in Fig.~\ref{fig:method 1}, which includes two steps(\textit{i.e.}, Generative Distillation and Attention Distillation). Given a video feature $F$, in the step 1, the teacher model leverages a pre-trained generative model, CVAE, to transfer the knowledge of feature and attention, which is to ensure the categories accuracy. In the step 2, the teacher would produce attention map $\lambda$, which represents the feature semantics of teacher model.

\subsubsection{Generative Distillation} 

In order to reconstruct the feature semantics, we introduce a \textit{Generative Distillation} module by transferring attention-based feature from teacher model. Furthermore, conditional variational autoencoder(CVAE) effectively reconstructs the feature semantic by learning the correspondence between feature map and attention map. In specific, as Fig.~\ref{fig:method 1}(a) shows, we freeze the student model and the attention module, update the CVAE module by leveraging the CVAE of teacher model. For updating CVAE, we build an action-based generative problem:
\begin{equation}
    p_\phi(F_t \vert \lambda_t) = \mathbb{E}_{p_\phi(z_t \vert \lambda_t)}[p_\phi(F_t \vert \lambda_t, z_t)],
\end{equation}
where $z_t$ is the latent variable, $\phi$ indicates the learnable model, $p_\phi(z_t \vert \lambda_t)$ indicates the prior model, and $p_\phi(F_t \vert \lambda_t, z_t)$ is the posterior model, which is the decoder procedure of the generative model. Notably, the latent variable $z_t$ is sampled from the learned prior distribution, which is set as the process of feature reconstruction.

During the training procedure, our target is to let the student model learn the reconstructed feature from teacher model, which is aiming at fitting the distribution of the teacher model and improving the the reconstructed quality of the student model. In specific, we propose the generative loss $\mathcal{L}_{KD-gen}$ to optimize the feature reconstruction:
\begin{equation}
    \mathcal{L}_{KD-gen} =\Vert f^T(F^T,A^T) - f^S(F^S,A^S) \Vert_2^2,
\end{equation}
where $S$ and $T$ indicate the student model and teacher model, respectively, $F$ represents the feature map, $A$ is the attention map and $f(\cdot)$ is the attention-based reconstructed feature function. Remarkably, we set the same input, the feature map and attention map of student, in the CVAE module of teacher and student model. The same input is to allow student to focus on the attention-based feature reconstruction of the teacher, without being affected by the differences in feature and attention maps extracted by the different model.

During the CVAE training, as Fig.~\ref{fig:method 1}(a) shows, our target is to minimize the evidence lower bound(ELBO) loss $\mathcal{L}_{cvae}$:
\begin{equation}
    \mathcal{L}_{cvae}=\log p_\psi(F_t|\lambda_t,z_t)+\alpha\cdot KL(q_\phi(z_t|F_t,\lambda_t)||p_\psi(z_t|\lambda_t)),
\end{equation}
where $z_t$ is the latent variable and $\alpha = 0.1$ is an empirical hyperparameter. In specific, as Fig.~\ref{fig:meth 2}(a) shows, firstly, we use Encoder to generate the intermediate variable $ z_t $, with function $ f_z $ by the feature $ F_t $ and the attention $ \lambda_t $, as follows:
\begin{equation}
     z = f_z\left(f_{fc_i}\left(\lambda_i, \theta_i\left(f_i\right)\right)\right),
\end{equation}
where $ f_{fc_i} $ is the liner model, and $ \theta_i $ is a $ 1\times1\times1 $ 3D convolution, aiming at concatenating $ \lambda_i $ for $ f_{z} $ by reducing the channels' number and adding learnable model parameters. $ f_z $ is a mean and variance function for the intermediate variable $ z $, which generates the latent variable $ Z $. Then, we use Decoder to generate the reconstructed feature map $ \widehat{F} $,  which leverages latent variable $ Z $ and attention $ \lambda $ , as follows:
\begin{equation}
  \widehat{F} = \theta_i^{\prime}\left(f_{fc_i}\left(Z,\lambda_i\right)\right),
\end{equation}
where $ \theta_i^{\prime} $ is a $ 1\times n \times n $( $ n $ is the kernel size) 3D deconvolution, which is aiming at generating the reconstructed feature map $ \widehat{F} $ by adding the number of channels. We use $ \mathcal{L}_{cvae} $ to measure the difference between $ \widehat{F} $ and $ F $.

\subsubsection{Attention Distillation} 

As Equ.~\ref{equ:att} shows, the Generative Distillation mainly optimizes the first term $\log p(C \vert F,\lambda)$, this section mainly focuses on the second term $\log p(\lambda|F)$. The attention represents the feature distribution of the video, so the student model can learn the representation ability of the teacher model by learning the attention map~\cite{quader2020weight,zagoruyko2016paying}. Specifically,  as Fig.~\ref{fig:method 1} (b) shown, we freeze the CVAE, update the student model and attention module by leveraging  attention distillation. For the video feature $F$, we use the attention $A$ to represent feature semantics, as follows
\begin{equation}
    A = Sigmod(GN(f_{Conv1D}(F))),
\end{equation}
where $GN$ indicates the group normalization. Furthermore, we calculate the attention-based feature $F^{\prime}$ as follows:
\begin{equation}
    F^{\prime}= \sigma \times (A \times \theta(F)),
\end{equation}
where $\theta(\cdot)$ indicates the 3D transposed convolution, $\sigma$ is the scaling factor, which is defined as:
\begin{equation}
    \sigma = \frac{\Vert F \Vert}{\Vert A \times \theta(F) \Vert}.
\end{equation}

As Fig.~\ref{fig:method 1} (b) shows, for distilling the attention, we propose the attention distillation loss $L_{KD-att}$, defined as:
\begin{equation}
    L_{KD-att} = \Vert A^T - A^S \Vert_2^2,
\end{equation}
where $A^T$ and $A^S$ indicate the attention map of the teacher model and the student model, respectively.

During the training, for action recognition, we encourage higher capability of the action classification. This amounts to minimizing the following loss:
\begin{equation}
   \mathcal{L}_{clf}= \sum_{c=1}^{C+1}-y_c(x)\log(p_c(x)),
\end{equation}
where $y_c(x)$ and $p_c(x)$ indicate action probability distribution of the ground truth and the predicted result respectively.

In the meanwhile, as shown in Fig.~\ref{fig:meth 2}(b), we use Decoder to generate reconstructed feature map, which is aiming at maintaining the distribution of attention-based feature during the feature representation. This amounts to minimizing the reconstruction loss $\mathcal{L}_{recon}$:
\begin{equation}
\label{con:r}
\begin{aligned}
  \mathcal{L}_{recon} &= -\log (\sum_{l=1}^L p_\psi\left(f_t \vert \lambda_t,z_t\right)) \\  
  &\ \ \ \ \ \backsimeq -\sum_{t=1}^T \log \left\{\frac{1}{L}\sum_{l=1}^L p_\psi\left(f_t\rvert\lambda_t,z_t^{\left(l\right)}\right)\right\} \\
  &\ \ \ \ \ \backsimeq  \Vert f_t - f_\psi\left(\lambda_t, z_t\right)\Vert^2,
\end{aligned}
\end{equation}
where $ z_t^{\left(l\right)} $ is generated by $z_t$ and $ \lambda_t $ in the encoder of generative model. Especially, following~\cite{shi2020weakly},  in the last step, we set $ L $ as 1.

\subsubsection{Training and Inference}

In our proposed KD framework, attention module requires attention-based feature as the input. Therefore, as shown in Fig.~\ref{fig:method 1}, we consider alternating the training of the attention module with the backbone module.

As Fig.~\ref{fig:method 1}(a) shows, in Stage 1, we freeze the student backbone and activate the CVAE module, which is to optimize the generative distillation procedure by the features extracted from the backbone and the generated attention. The overall loss function of the stage 1 is as follows:
\begin{equation}
    \mathcal{L}_{GD} = \mathcal{L}_{CVAE} + \beta\cdot\mathcal{L}_{KD-gen},
\end{equation}
where $\beta$ is  an empirical hyperparameter, $\beta = 0.01$.

As Fig.~\ref{fig:method 1}(b) shows, in Stage 2, we freeze the CVAE module and activate the student backbone, which is to optimize the attention distillation procedure by the attention-based reconstructed feature, the attention distribution and the predicted action probability, as follows:
\begin{equation}
    \mathcal{L}_{AD} = \mathcal{L}_{recon} + \mathcal{L}_{clf} + \gamma \cdot \mathcal{L}_{KD-att},
\end{equation}
where $\gamma$ is  an empirical hyperparameter, $\gamma = 0.1$. Notably, during the inference procedure, we only exploit student model with attention module to predict action label, not including generative model.

\section{Experiments}\label{sec:Experiment}
In this section, we first describe datasets and evaluation metrics. Then, we evaluate our model's effectiveness followed by main result and ablation study.

\subsection{Datasets and Evaluation Metrics}
To validate the effectiveness of our model, we conduct extensive experiments on commonly-used action recognition benchmark UCF101~\cite{soomro2012ucf101} and HMDB51~\cite{kuehne2011hmdb}, commonly-used action detection  benchmark THUMOS14~\cite{THUMOS14}.

\subsubsection{UCF101} It consists of 13320 action videos, including 101 action categories, which has 3 official splits and each split divides the training set and test set at a ratio of 7:3.

\subsubsection{HMDB51} It consists of 6849 video clips, which contains 51 action categories and each category includes at least 101 video clips. It has the same split ratio with UCF101 dataset.

\subsubsection{THUMOS14} It contains 101 categories of videos and is composed of four parts: training, validation, testing and background set. Each set includes 13320, 1010, 1574 and 2500 videos, respectively. Following the common setting~\cite{THUMOS14}, we used 200 videos in the validation set for training, 213 videos in the testing set for evaluation.

\subsubsection{Evaluation Metrics} We follow the standard evaluation protocol and report accuracy as evaluation metric. In Specific, we report the Top-1 and Top-5 on action recognition. We report the mean Average Precision(mAP) at the different intersections over union(IoU) thresholds on action detection. 

\subsection{Implementation Details}

On UCF101 and HMDB51, we use tvl1~\cite{brox2009large} to extract optical frames. The length of the clip is set to $64$. We resize the frame to $256$ for UCF101 and $240$ for HMDB51. We use SGD optimizer for student model and Adam optimizer for KD module. 


On THUMOS14, we sample RGB and optical flow at 10 fps and split video into clips of 256 frames. Adjacent clips have a temporal overlap of 30 frames during training and 128 frames during testing. The size of frame is set to $96\times96$.


We adopt offline distillation strategy and transferred knowledge by alternately training generative distillation module and attention distillation module.

\subsection{Main Results}

\begin{table*}[htbp]\centering
\caption{Validation accuracy and computation cost on UCF101 and HMDB51. We set I3D as the teacher model, Top-I3D as the student model. For fair comparison, we keep the same training configuration for all methods.}
\label{tab1}
\setlength{\tabcolsep}{3.3mm}{
    \begin{tabular}{ccccccc}
    \toprule
     \multirow{2}*{Model}  &   \multirow{2}*{Knowledge} &   \multicolumn{2}{c}{UCF101}   &   \multicolumn{2}{c}{HMDB51}  &   \multirow{2}*{FLOPS(G)}\\
     \cmidrule(lr){3-4}\cmidrule(lr){5-6}
      & & Top-1   &   Top-5   &   Top-1   &   Top-5\\
     \midrule
     Teacher    &   -  &   91.9    &   98.8    &   69.0    &   88.8    &   111.3\\
     Student    &   -  &   64.1    &   82.1    &   52.0    &   77.6    &   45.5\\
     KD~\cite{hinton2015distilling}   &   Logits  &   65.2    &   79.2    &   53.1    &   78.2    &   45.5\\
     CTKD~\cite{li2023curriculum}   &   Logits  &   65.6   &    83.5   &    52.6   &    78.1   &   45.5\\
     FN~\cite{xu2020feature} &   Feature  &   65.5    &   79.4    &   52.9    &   77.3    &   45.5\\
     MGD~\cite{yang2022masked}  &   Feature &   65.4   &    79.9   &    54.1   &    77.9   &   45.5\\
     SimKD~\cite{chen2022knowledge} &   Feature &   66.1   &    81.0   &    53.6   &    78.9   &   45.5\\
     AT~\cite{zagoruyko2016paying} &   Attention   &   66.2    &   80.5    &   52.9    &   77.1    &   47.4\\
     CTKD~\cite{zhao2020highlight}    &   Logits+Feature   &   65.7    &   80.1    &   53.9    &   78.8    &   45.5\\
     Ours   &   Feature   &   \textbf{66.6}    &   \textbf{84.5}    &   \textbf{54.5}    &   \textbf{79.0}    &   47.4\\
     \bottomrule
\end{tabular}}
\end{table*}

To verify the effectiveness of our method, we compare our method with other distillation methods on the action recognition and the action detection task. In the Table~\ref{tab1}, we adopt the I3D~\cite{carreira2017quo} as the teacher model and the Top-I3D~\cite{xie2018rethinking} as the student model. In the Table~\ref{tab2},  we adopt the AFSD~\cite{lin2021learning}as the teacher model, which is the CNN-based sate-of-the-art method on the action detection, and the Top-I3D-based AFSD as the student model. The Top-I3D replaces partial 3D convolutional blocks of the I3D network with 2D convolutional blocks, which reduces the parameters and computation cost. 

As shown in Table~\ref{tab1}, for the different datasets of action recognition task, our method outperforms the previous knowledge distillation methods and achieves significant improvement. To be specific, on the UCF101, based on the student model, our method gains a significant increase of $2.5\%$ and $2.4\%$ on the Top-1 and the Top-5 accuracy, which finally reaches $66.6\%$ and $84.5\%$, respectively. On the more complex dataset HMDB51, based on the student model, our method also achieves $2.5\%$ and $1.4\%$ increase, which reaches $54.5\%$ and $79.0\%$. 

For fairness, we adopt the same backbone (I3D-based teacher and Top-I3D-based student) with other knowledge distillation methods. In particular, our approach outperforms other feature-based KD methods (\textit{e.g.}, FN~\cite{xu2020feature}, SimKD~\cite{chen2022knowledge}), which also utilize the feature distillation to transfer knowledge but without modeling critical area. Furthermore, our approach outperforms AT~\cite{zagoruyko2016paying}, which also utilizes the attention distillation but without explicit feature semantics. The results demonstrate the superior performance of our approach with attention-based feature semantics modeling.

\begin{table*}[htbp]\centering
\caption{Validation accuracy and computation cost on Thumos14. The teacher model is AFSD, student model is Top-I3D-based AFSD. We keep the same configuration for all methods.}
\label{tab2}
\setlength{\tabcolsep}{4.0mm}{
    \begin{tabular}{cccccccc}
    \toprule
    mAP(\%)/tIOU & 0.1 & 0.2 & 0.3 & 0.4 & 0.5 & AVG & FLOPS(G) \\
    \midrule
    Teacher & 71.8 & 70.6 & 67.3 & 62.4 & 55.5 & 65.5 & 84.4\\
    Student & 37.9 & 36.7 & 33.1 & 30.1 & 25.2 & 32.6 & 36.0\\
    KD~\cite{hinton2015distilling}  &   38.5    &   37.0    &   34.3    &   31.0    &   24.6    &   33.1    &   36.0\\
    SimKD~\cite{chen2022knowledge} & 38.5 & 37.1 & 34.6 &30.9 &25.1 & 33.2 & 36.0\\
    Ours    & \textbf{39.4} & \textbf{37.5} & \textbf{34.9} & \textbf{31.5} & \textbf{25.3} & \textbf{33.7} & 39.9\\
    \bottomrule
\end{tabular}}
\end{table*}

In addition to experimenting on different datasets, we expanded our knowledge distillation method on different task. We conduct experiment on the Thumos14 dataset for action detection task and the comparison results are summarized in Table~\ref{tab2}. Again, based on the student model, our proposed method obtains a significant improvement of $1.1\%$ average mAP, surpassing the other works(\textit{e.g.} KD~\cite{hinton2015distilling}). The consistent superior results on different tasks justify the effectiveness of our proposed method. As shown in Table~\ref{tab2}, on the more complex task, action detection, our method achieves improvement in all tIOU $[0.1:0.1:0.5]$ and reaches the $33.7\%$ average mAP.

In contrast to the teacher model, as shown in Tables~\ref{tab1} and~\ref{tab2}, the student model demonstrates a discernible decrease in accuracy across different video tasks. Nonetheless, it is noteworthy that the computational expense associated with the student amounts to merely approximately $40.9\%$ of that of the teacher. Concurrently, with the help of our framework, the student model gains considerable improvement, with a marginal increment in computational overhead.

\subsection{Ablation Study}
To demonstrate the reasonableness of our framework, we analyze the effect of every submodule and some function operations in this subsection.
\subsubsection{Contribution of each sub-module} 
We study each sub-module influence on the overall performance of action recognition. We start from the baseline which only includes classification loss. Next we add feature representation loss $L_{recon}$, note that it involves both attention module and attention-based feature representation module. We then introduce the generative distillation loss $L_{KD-cvae}$ and attention distillation loss $L_{KD-att}$, respectively.

\begin{table}[htbp]\centering
\caption{Performance of each sub-module on UCF101.}
\label{tab3}
    \setlength{\tabcolsep}{1.2mm}{
    \begin{tabular}{ccccccc}
    \toprule
    \textit{baseline}& $L_{recon}$ & $L_{KD-cvae}$ & $L_{KD-att}$ & Top-1 & Top-5\\
    \midrule
    $\checkmark$ & & &              & 64.1 & 82.1\\
    $\checkmark$  & $\checkmark$  &   &   &   65.2    &   83.3\\
   $\checkmark$  &  $\checkmark$  &$\checkmark$  &              & 66.2 & 84.0\\
    $\checkmark$  & $\checkmark$  &              & $\checkmark$ & 66.1 & 83.5\\
    $\checkmark$  & $\checkmark$  & $\checkmark$ & $\checkmark$ & \textbf{66.6} & \textbf{84.5}\\
    \bottomrule
\end{tabular}}
\end{table}

Table~\ref{tab3} summarizes the performances by considering one more factor at each stage on UCF101. We first observe that adding the feature representation loss $L_{recon}$ and the KD loss $L_{KD-gen}$,$L_{KD-att}$ largely enhance the performance of the action recognition model. The $L_{recon}$ is aiming at generating attention to represent feature semantics, which also improves the ability of feature representation. The two distillation losses encourage the feature learning in the student model by transferring the attention knowledge and the feature knowledge. As shown in Table~\ref{tab3}, based on the baseline, our feature representation contributes an increase $1.1\%$ , $1.2\%$ and reaches $65.2\%$, $83.3\%$ on the Top-1 and Top-5 respectively. Based on this, our two distillation losses brings $2.1\%$, $1.9\%$ and $2.0\%$, $1.4\%$ on the Top-1 and Top-5 respectively, which finally reach $66.2\%$, $84.0\%$ and $66.1\%$, $83.5\%$. When introducing both distillation losses simultaneously, compared with the baseline, our proposed method improve $2.5\%$ and $2.4\%$ on the Top-1 and Top-5 accuracy.

\subsubsection{Effectiveness of attention module}
For transferring the feature semantics, We distill the attention by representing spatial-temporal feature. In specific, during the procedure of generating the attention, we leverage the generative model CVAE to optimize the attention module, which is aiming at improving the accuracy of the model and learning the feature information. To see this, we conduct the experiment by directly adding our attention-based feature representation module into different model, I3D~\cite{carreira2017quo} and Top-I3D~\cite{xie2018rethinking}.

As shown in Table~\ref{tab4}, compared to the I3D, our attention representation module contributes an increase of $0.5\%$, $0.3\%$ and the performance finally reaches $92.4\%$, $99.1\%$ on the Top1 and the Top5. Compared to the Top-I3D model, our attention representation module contributes an increase of $0.6\%$, $0.8\%$ and the performance finally reaches $64.7\%$, $82.9\%$ on the Top1 and the Top-5. Both results show that adding our attention representation module can represent the feature semantics and improve the model performance. 
\begin{table}[htbp]\centering
\caption{Performance of attention-based semantic representation module on different student models on UCF101.}
\label{tab4}
    \setlength{\tabcolsep}{5mm}{
    \begin{tabular}{ccc}
    \toprule
     Model          &   Top-1   &   Top-5\\
     \midrule
     I3D            &   91.9    &   98.8\\
     I3D+Att        &   \textbf{92.4 }   &   \textbf{99.1}\\
     Top-I3D        &   64.1    &   82.1\\
     Top-I3D+Att    &   \textbf{64.7}    &   \textbf{82.9}\\
     \bottomrule
\end{tabular}}
\end{table}

\subsubsection{Generalization of our method}
To verify the generalization of our proposed KD framework, we expand the experiment on different student model~\cite{xie2018rethinking}(\textit{i.e.}, Top-I3D, Bottom-I3D and I2D). As shown in Table~\ref{tab5}, our method is effective to all different student models. Compared to the Top-I3D model, our KD framework contributes an increase of $2.1\%$, $2.8\%$ and the performance finally reaches $66.2\%$, $84.9\%$ on the Top1 and the Top5. Compared to the Bottom-I3D model, our KD framework contributes an increase of $0.6\%$, $0.5\%$ and the performance finally reaches $53.3\%$, $72.9\%$ on the Top1 and the Top5. In particular, in the experiment with I2D, which is a 2D-based convolutional neural network, our method boosts the Top-1 accuracy by $6.4\%$, the Top-5 accuracy by $5.2\%$ and the performance finally reaches $58.6\%$, $76.0\%$, which even achieves comparable performance with 3D convolutional neural network. 

\begin{table}[htbp]\centering
\caption{Performance of our proposed framework on different student model on UCF101.}
\label{tab5}
    \setlength{\tabcolsep}{5mm}{
    \begin{tabular}{ccc}
     \toprule
     Model         & Top-1 & Top-5\\
     \midrule
     Top-I3D       & 64.1  & 82.1\\
     Top-I3D+Ours    & \textbf{66.2}  & \textbf{84.9}\\
     Bottom-I3D    & 52.7  & 72.7\\
     Bottom-I3D+Ours & \textbf{53.3}  & \textbf{72.9}\\
     I2D           & 52.2  & 70.8\\
     I2D+Ours        & \textbf{58.6}  & \textbf{76.0}\\
     \bottomrule
\end{tabular}}
\end{table}

\begin{figure}[h]
  \centering
  \begin{minipage}[t]{0.15\textwidth}
  \centering
  \includegraphics[width=\textwidth]{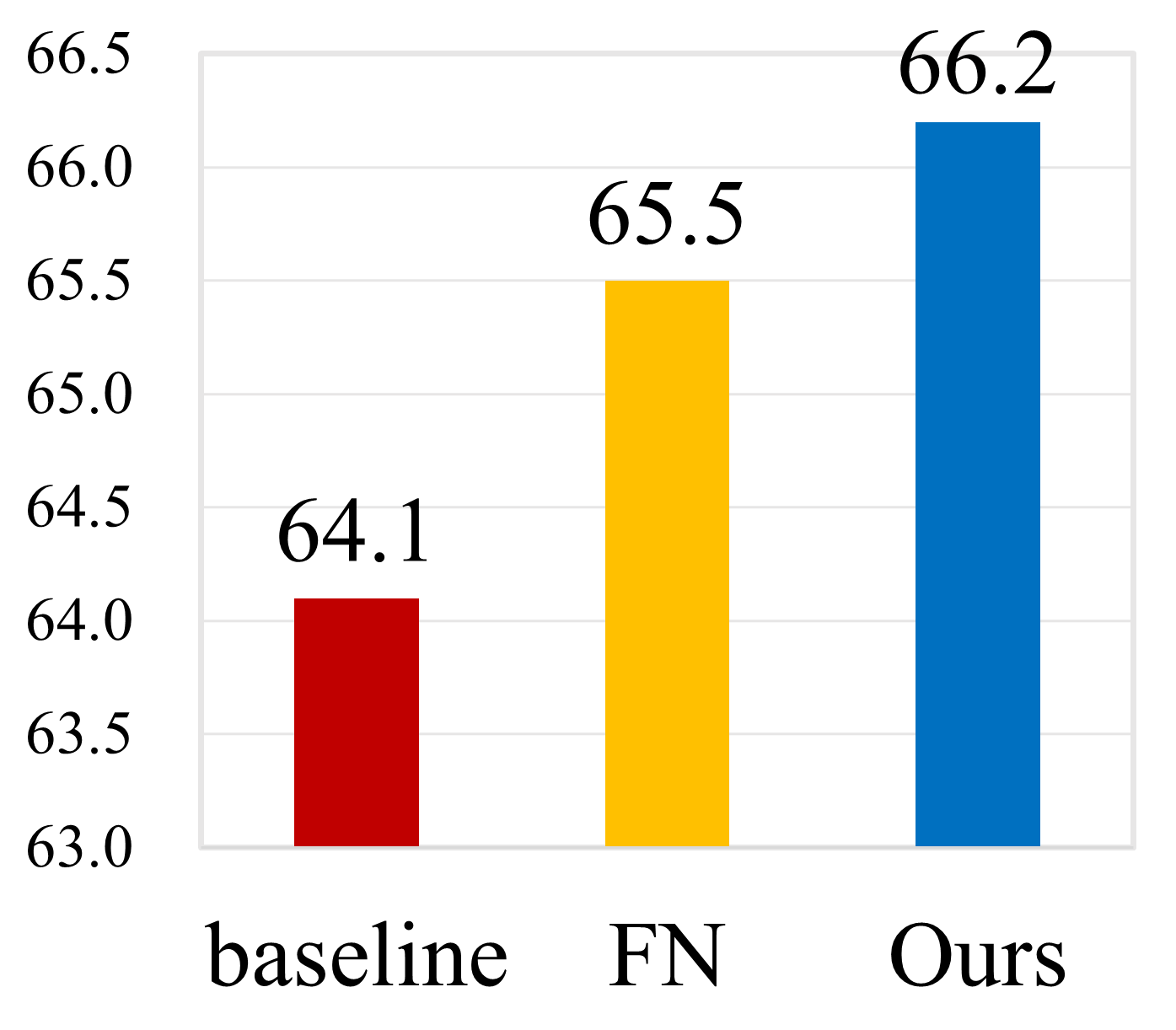}
  \caption*{Top-I3D}
  \end{minipage}
  \centering
  \begin{minipage}[t]{0.15\textwidth}
  \centering
  \includegraphics[width=\textwidth]{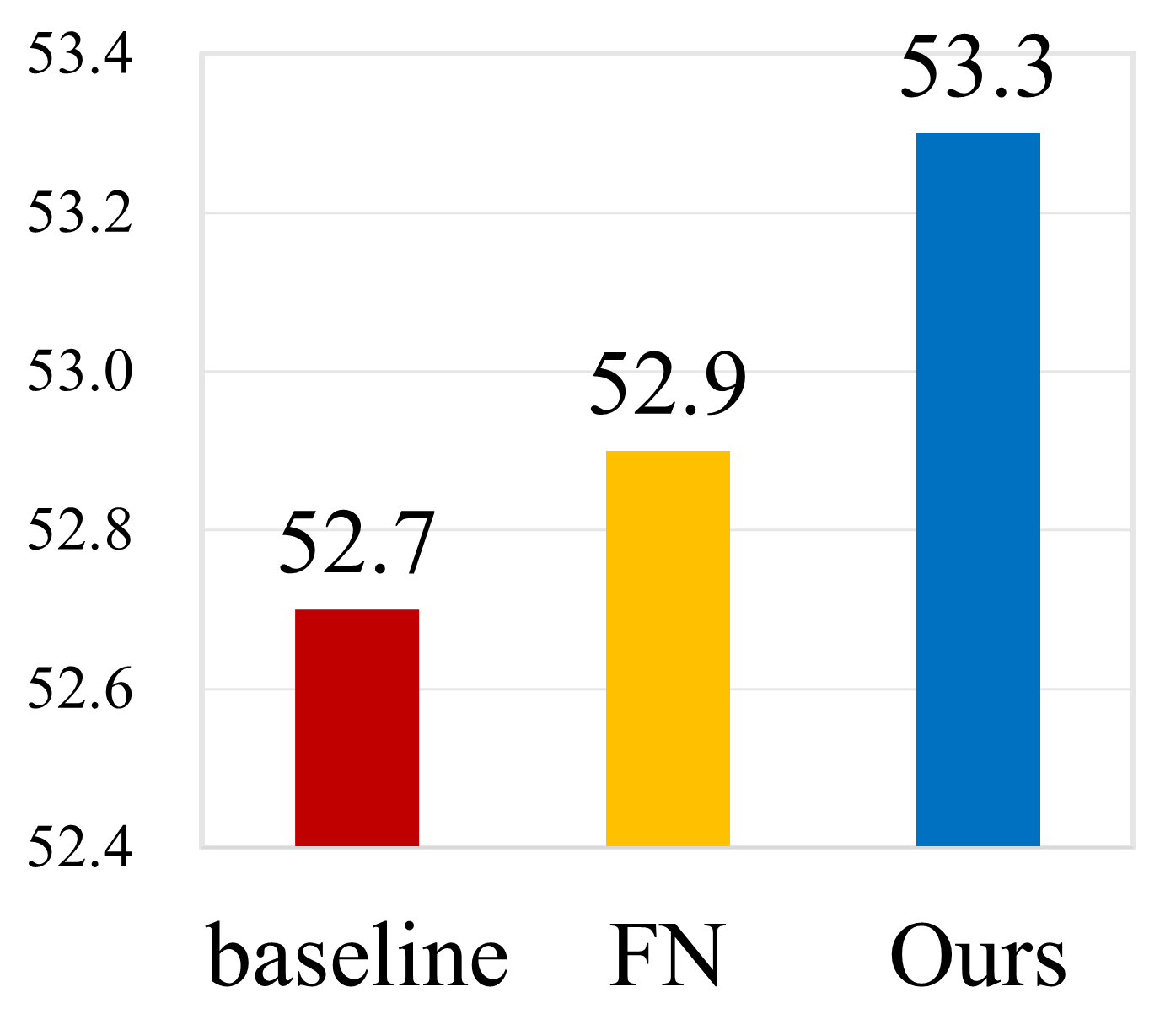}
  \caption*{Bottom-I3D}
  \end{minipage}
  \centering
  \begin{minipage}[t]{0.15\textwidth}
  \centering
  \includegraphics[width=\textwidth]{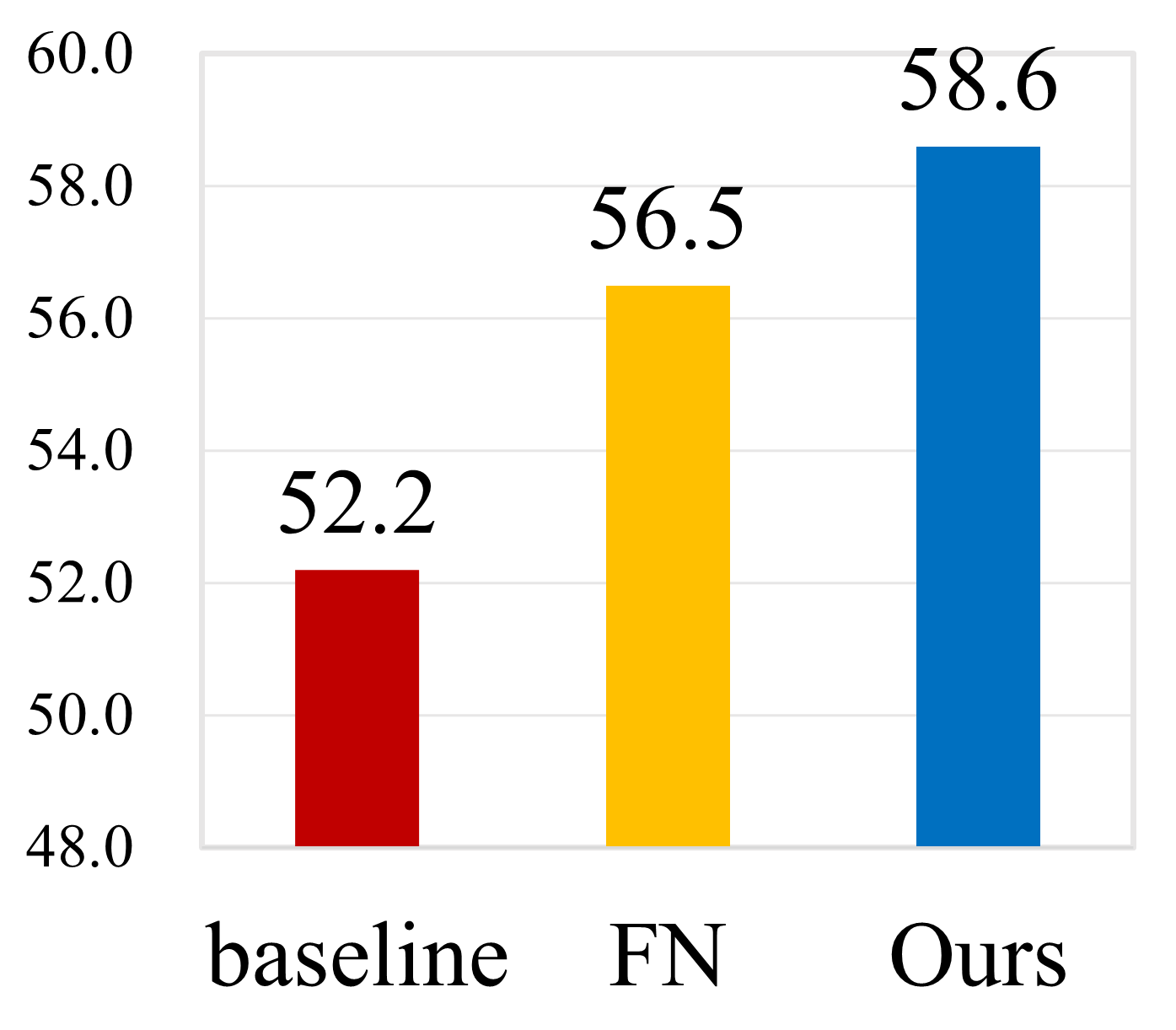}
  \caption*{I2D}
  \end{minipage}
  \caption{Accuracy of our proposed framework on different student models with FN on UCF101.}
  \label{fig:exp 1}
\end{figure}

Furthermore, we present a comparative evaluation of our method against the feature-based approach proposed by FN~\cite{xu2020feature}, thereby demonstrating the enhanced advancements and efficiency achieved by our proposed method.
As Fig.~\ref{fig:exp 1} shows, our method achieves better performance improvement ($1.4\%$ vs $2.1\%$ on Top-I3D, $0.2\%$ vs $0.6\%$ on Bottom-I3D, $4.3\%$ vs $6.4\%$ on I2D) on different student models. 
Significantly, our method demonstrates a noteworthy augmentation of $6.4\%$ on the I2D, leading to a final performance level of $58.6\%$, which proves the efficacy of our approach in distilling knowledge from 3D to 2D-CNNs.

\section{Conclusions}\label{sec:Conclusion}

In this paper, we propose a novel generative model-based knowledge distillation framework to transfer video feature semantics through attention representation. Our knowledge distillation framework mainly encompasses two primary steps: Feature Representation and Generative-based Distillation (Generative Distillation and Attention Distillation). By leveraging the attention mechanisms, we capture feature semantics and effectuate knowledge distillation through generative modeling. In the realm of 3D-CNNs distillation, our method exhibits a remarkable performance advancement over existing approaches across diverse video action analysis tasks. The results demonstrate the efficacy of our approach in enhancing knowledge distillation performance via generative model-based feature semantic transfer. Consequently, this study establishes the groundwork for an innovative knowledge distillation framework with a specific focus on 3D model distillation. For the future work, we envision the generative model paradigm as a promising direction for various video knowledge distillation tasks. It is also worth to explore such mechanism in other related tasks.


\section{Acknowledgements}\label{sec:ack}
This work was supported in part by the National Key Research and Development Program of China under Grants 2021YFB2401300, 2022YFA1004100,
and 2020YFA0713900; and in part by the National Natural Science Foundation of China under Grants 62172329, U1811461, U21A6005, and 11690011
\bibliography{aaai23}

\begin{thebibliography}{39}
\providecommand{\natexlab}[1]{#1}

\bibitem[{Brox, Bregler, and Malik(2009)}]{brox2009large}
Brox, T.; Bregler, C.; and Malik, J. 2009.
\newblock Large displacement optical flow.
\newblock In \emph{CVPR}.

\bibitem[{Carreira and Zisserman(2017)}]{carreira2017quo}
Carreira, J.; and Zisserman, A. 2017.
\newblock Quo vadis, action recognition? a new model and the kinetics dataset.
\newblock In \emph{CVPR}.

\bibitem[{Chen et~al.(2022)Chen, Mei, Zhang, Wang, Feng, and
  Chen}]{chen2022knowledge}
Chen, D.; Mei, J.-P.; Zhang, H.; Wang, C.; Feng, Y.; and Chen, C. 2022.
\newblock Knowledge distillation with the reused teacher classifier.
\newblock In \emph{CVPR}.

\bibitem[{Crasto et~al.(2019)Crasto, Weinzaepfel, Alahari, and
  Schmid}]{crasto2019mars}
Crasto, N.; Weinzaepfel, P.; Alahari, K.; and Schmid, C. 2019.
\newblock Mars: Motion-augmented rgb stream for action recognition.
\newblock In \emph{CVPR}.

\bibitem[{Dai, Das, and Bremond(2021)}]{dai2021learning}
Dai, R.; Das, S.; and Bremond, F. 2021.
\newblock Learning an augmented rgb representation with cross-modal knowledge
  distillation for action detection.
\newblock In \emph{ICCV}.

\bibitem[{Foo et~al.(2023)Foo, Gong, Fan, and Liu}]{foo2023system}
Foo, L.~G.; Gong, J.; Fan, Z.; and Liu, J. 2023.
\newblock System-status-aware Adaptive Network for Online Streaming Video
  Understanding.
\newblock In \emph{CVPR}.

\bibitem[{Goodfellow et~al.(2014)Goodfellow, Pouget-Abadie, Mirza, Xu,
  Warde-Farley, Ozair, Courville, and Bengio}]{goodfellow2014generative}
Goodfellow, I.; Pouget-Abadie, J.; Mirza, M.; Xu, B.; Warde-Farley, D.; Ozair,
  S.; Courville, A.; and Bengio, Y. 2014.
\newblock Generative adversarial nets.
\newblock \emph{NeurIPS}.

\bibitem[{Hara, Kataoka, and Satoh(2017)}]{hara2017learning}
Hara, K.; Kataoka, H.; and Satoh, Y. 2017.
\newblock Learning spatio-temporal features with 3d residual networks for
  action recognition.
\newblock In \emph{ICCV}.

\bibitem[{Hinton, Vinyals, and Dean(2015)}]{hinton2015distilling}
Hinton, G.; Vinyals, O.; and Dean, J. 2015.
\newblock Distilling the knowledge in a neural network.
\newblock \emph{arXiv preprint arXiv:1503.02531}.

\bibitem[{Ji et~al.(2012)Ji, Xu, Yang, and Yu}]{ji20123d}
Ji, S.; Xu, W.; Yang, M.; and Yu, K. 2012.
\newblock 3D convolutional neural networks for human action recognition.
\newblock \emph{TPAMI}.

\bibitem[{Jiang et~al.(2014)Jiang, Liu, Roshan~Zamir, Toderici, Laptev, Shah,
  and Sukthankar}]{THUMOS14}
Jiang, Y.-G.; Liu, J.; Roshan~Zamir, A.; Toderici, G.; Laptev, I.; Shah, M.;
  and Sukthankar, R. 2014.
\newblock {THUMOS} Challenge: Action Recognition with a Large Number of
  Classes.
\newblock \url{http://crcv.ucf.edu/THUMOS14/}.

\bibitem[{Kingma and Welling(2014)}]{kingma2013auto}
Kingma, D.~P.; and Welling, M. 2014.
\newblock Auto-Encoding Variational Bayes.
\newblock In \emph{ICLR}.

\bibitem[{Kuehne et~al.(2011)Kuehne, Jhuang, Garrote, Poggio, and
  Serre}]{kuehne2011hmdb}
Kuehne, H.; Jhuang, H.; Garrote, E.; Poggio, T.; and Serre, T. 2011.
\newblock HMDB: a large video database for human motion recognition.
\newblock In \emph{ICCV}.

\bibitem[{Li et~al.(2023)Li, Li, Yang, Zhao, Song, Luo, Li, and
  Yang}]{li2023curriculum}
Li, Z.; Li, X.; Yang, L.; Zhao, B.; Song, R.; Luo, L.; Li, J.; and Yang, J.
  2023.
\newblock Curriculum temperature for knowledge distillation.
\newblock In \emph{AAAI}.

\bibitem[{Lin et~al.(2021)Lin, Xu, Luo, Wang, Tai, Wang, Li, Huang, and
  Fu}]{lin2021learning}
Lin, C.; Xu, C.; Luo, D.; Wang, Y.; Tai, Y.; Wang, C.; Li, J.; Huang, F.; and
  Fu, Y. 2021.
\newblock Learning salient boundary feature for anchor-free temporal action
  localization.
\newblock In \emph{CVPR}.

\bibitem[{Lin et~al.(2022)Lin, Xie, Wang, Yu, Chang, Liang, and
  Wang}]{Lin_2022_CVPR}
Lin, S.; Xie, H.; Wang, B.; Yu, K.; Chang, X.; Liang, X.; and Wang, G. 2022.
\newblock Knowledge Distillation via the Target-Aware Transformer.
\newblock In \emph{CVPR}.

\bibitem[{Liu et~al.(2021)Liu, Wang, Li, and Lin}]{liu2021semantics}
Liu, Y.; Wang, K.; Li, G.; and Lin, L. 2021.
\newblock Semantics-aware adaptive knowledge distillation for sensor-to-vision
  action recognition.
\newblock \emph{TIP}.

\bibitem[{Qiu, Yao, and Mei(2017)}]{qiu2017learning}
Qiu, Z.; Yao, T.; and Mei, T. 2017.
\newblock Learning spatio-temporal representation with pseudo-3d residual
  networks.
\newblock In \emph{ICCV}.

\bibitem[{Quader et~al.(2020)Quader, Bhuiyan, Lu, Dai, and
  Li}]{quader2020weight}
Quader, N.; Bhuiyan, M. M.~I.; Lu, J.; Dai, P.; and Li, W. 2020.
\newblock Weight excitation: Built-in attention mechanisms in convolutional
  neural networks.
\newblock In \emph{ECCV}.

\bibitem[{Romero et~al.(2015)Romero, Ballas, Kahou, Chassang, Gatta, and
  Bengio}]{romero2014fitnets}
Romero, A.; Ballas, N.; Kahou, S.~E.; Chassang, A.; Gatta, C.; and Bengio, Y.
  2015.
\newblock FitNets: Hints for Thin Deep Nets.
\newblock In \emph{ICLR}.

\bibitem[{Shen et~al.(2019)Shen, Xue, Wang, Song, Sun, and
  Song}]{shen2019customizing}
Shen, C.; Xue, M.; Wang, X.; Song, J.; Sun, L.; and Song, M. 2019.
\newblock Customizing student networks from heterogeneous teachers via adaptive
  knowledge amalgamation.
\newblock In \emph{ICCV}.

\bibitem[{Shi et~al.(2020)Shi, Dai, Mu, and Wang}]{shi2020weakly}
Shi, B.; Dai, Q.; Mu, Y.; and Wang, J. 2020.
\newblock Weakly-supervised action localization by generative attention
  modeling.
\newblock In \emph{CVPR}.

\bibitem[{Sohn, Lee, and Yan(2015)}]{sohn2015learning}
Sohn, K.; Lee, H.; and Yan, X. 2015.
\newblock Learning structured output representation using deep conditional
  generative models.
\newblock \emph{NeurIPS}.

\bibitem[{Soomro, Zamir, and Shah(2012)}]{soomro2012ucf101}
Soomro, K.; Zamir, A.~R.; and Shah, M. 2012.
\newblock UCF101: A dataset of 101 human actions classes from videos in the
  wild.
\newblock \emph{arXiv preprint arXiv:1212.0402}.

\bibitem[{Stroud et~al.(2020)Stroud, Ross, Sun, Deng, and
  Sukthankar}]{stroud2020d3d}
Stroud, J.; Ross, D.; Sun, C.; Deng, J.; and Sukthankar, R. 2020.
\newblock D3d: Distilled 3d networks for video action recognition.
\newblock In \emph{WACV}.

\bibitem[{Sun et~al.(2022)Sun, Ke, Rahmani, Bennamoun, Wang, and
  Liu}]{sun2022human}
Sun, Z.; Ke, Q.; Rahmani, H.; Bennamoun, M.; Wang, G.; and Liu, J. 2022.
\newblock Human action recognition from various data modalities: A review.
\newblock \emph{TPAMI}.

\bibitem[{Thoker and Gall(2019)}]{thoker2019cross}
Thoker, F.~M.; and Gall, J. 2019.
\newblock Cross-modal knowledge distillation for action recognition.
\newblock In \emph{ICIP}.

\bibitem[{Tran et~al.(2015)Tran, Bourdev, Fergus, Torresani, and
  Paluri}]{tran2015learning}
Tran, D.; Bourdev, L.; Fergus, R.; Torresani, L.; and Paluri, M. 2015.
\newblock Learning spatiotemporal features with 3d convolutional networks.
\newblock In \emph{ICCV}.

\bibitem[{Tran et~al.(2018)Tran, Wang, Torresani, Ray, LeCun, and
  Paluri}]{tran2018closer}
Tran, D.; Wang, H.; Torresani, L.; Ray, J.; LeCun, Y.; and Paluri, M. 2018.
\newblock A closer look at spatiotemporal convolutions for action recognition.
\newblock In \emph{CVPR}.

\bibitem[{Xie et~al.(2018)Xie, Sun, Huang, Tu, and Murphy}]{xie2018rethinking}
Xie, S.; Sun, C.; Huang, J.; Tu, Z.; and Murphy, K. 2018.
\newblock Rethinking spatiotemporal feature learning: Speed-accuracy trade-offs
  in video classification.
\newblock In \emph{ECCV}.

\bibitem[{Xu, Das, and Saenko(2017)}]{xu2017r}
Xu, H.; Das, A.; and Saenko, K. 2017.
\newblock R-c3d: Region convolutional 3d network for temporal activity
  detection.
\newblock In \emph{ICCV}.

\bibitem[{Xu et~al.(2020)Xu, Rui, Li, and Gu}]{xu2020feature}
Xu, K.; Rui, L.; Li, Y.; and Gu, L. 2020.
\newblock Feature normalized knowledge distillation for image classification.
\newblock In \emph{ECCV}.

\bibitem[{Yang et~al.(2022{\natexlab{a}})Yang, Zhang, Xu, Yu, Fan, and
  Xu}]{yang2022massive}
Yang, S.; Zhang, L.; Xu, C.; Yu, H.; Fan, J.; and Xu, Z. 2022{\natexlab{a}}.
\newblock Massive data clustering by multi-scale psychological observations.
\newblock \emph{National Science Review}.

\bibitem[{Yang et~al.(2022{\natexlab{b}})Yang, Li, Jiang, Gong, Yuan, Zhao, and
  Yuan}]{yang2022focal}
Yang, Z.; Li, Z.; Jiang, X.; Gong, Y.; Yuan, Z.; Zhao, D.; and Yuan, C.
  2022{\natexlab{b}}.
\newblock Focal and global knowledge distillation for detectors.
\newblock In \emph{CVPR}.

\bibitem[{Yang et~al.(2022{\natexlab{c}})Yang, Li, Shao, Shi, Yuan, and
  Yuan}]{yang2022masked}
Yang, Z.; Li, Z.; Shao, M.; Shi, D.; Yuan, Z.; and Yuan, C. 2022{\natexlab{c}}.
\newblock Masked generative distillation.
\newblock In \emph{ECCV}.

\bibitem[{Zagoruyko and Komodakis(2017)}]{zagoruyko2016paying}
Zagoruyko, S.; and Komodakis, N. 2017.
\newblock Paying More Attention to Attention: Improving the Performance of
  Convolutional Neural Networks via Attention Transfer.
\newblock In \emph{ICLR}.

\bibitem[{Zhang and Ma(2020)}]{zhang2020improve}
Zhang, L.; and Ma, K. 2020.
\newblock Improve object detection with feature-based knowledge distillation:
  Towards accurate and efficient detectors.
\newblock In \emph{ICLR}.

\bibitem[{Zhao et~al.(2022)Zhao, Cui, Song, Qiu, and Liang}]{zhao2022decoupled}
Zhao, B.; Cui, Q.; Song, R.; Qiu, Y.; and Liang, J. 2022.
\newblock Decoupled knowledge distillation.
\newblock In \emph{CVPR}.

\bibitem[{Zhao et~al.(2020)Zhao, Sun, Dong, Chen, and Dong}]{zhao2020highlight}
Zhao, H.; Sun, X.; Dong, J.; Chen, C.; and Dong, Z. 2020.
\newblock Highlight every step: Knowledge distillation via collaborative
  teaching.
\newblock \emph{IEEE Transactions on Cybernetics}.

\end{thebibliography}

\end{document}